\documentclass[twoside,11pt]{article}

\usepackage{blindtext}

\usepackage{formatting}
\usepackage{float}
\usepackage{cleveref}
\usepackage{parskip}
\usepackage{microtype}
\usepackage{xcolor}
\usepackage{enumitem}
\hypersetup{colorlinks=true, allcolors=blue!30!black}
\usepackage[symbol]{footmisc}

\usepackage{lastpage}

\firstpageno{1}

\begin{document}

\title{A Library for Learning Neural Operators}

 \author{\name Jean Kossaifi\addr \(^{1*}\), \name Nikola Kovachki\addr \(^{1*}\), \name Zongyi Li\addr \(^{2*}\), \name David Pitt\addr \(^{2*}\), \name Miguel Liu-Schiaffini\addr \(^{2}\), \\ \name Robert J. George\addr \(^{2}\), \name Boris Bonev\addr \(^{1}\), \name Kamyar Azizzadenesheli\addr \(^{1}\), \name Julius Berner\addr \(^{1,2}\), \\ \name Valentin Duruisseaux\addr \(^{2}\) \and \name Anima Anandkumar\addr \(^{2}\)
\\
\addr \(^1\)NVIDIA \hfill\hfill \addr \(^2\)Caltech
}
\footnotetext[1]{These authors contributed equally to this work.}
\def\thefootnote{\arabic{footnote}}

\editor{} %

\maketitle

\begin{abstract}%
We present \textsc{NeuralOperator}, an open-source Python library for operator learning. Neural operators generalize neural networks to maps between function spaces instead of finite-dimensional Euclidean spaces. They can be trained and inferenced on input and output functions given at various discretizations, satisfying a discretization convergence properties. 
Part of the official PyTorch Ecosystem, \textsc{NeuralOperator} provides all the tools for training and deploying neural operator models, as well as developing new ones, in a high-quality, tested, open-source package.
It combines cutting-edge models and customizability with a gentle learning curve and simple user interface for newcomers.
\end{abstract}

\section{Introduction}
Most scientific problems involve mappings between functions, not finite-dimensional data: notably, partial differential equations (PDEs) are naturally described on function spaces. A practical use case would be, for instance, learning a solution operator mapping between initial conditions and solution functions.
Traditional numerical methods operate on discretizations of functions based on meshes of the computational domains, with their accuracy heavily depending on the meshes' resolutions. 
In concrete applications, such as weather or climate simulations, the requirement of a fine mesh renders such methods computationally intensive, making it intractable to simulate solutions across large sets of parameters (such as initial conditions or coefficients) in a reasonable amount of time~\citep{schneider2017climate}. 

Deep neural networks have been considered to accelerate the solution of PDEs by mapping from parameters directly to the solution on a given discretization. However, they can only learn mappings between finite-dimensional spaces, not function spaces.
In other words, the solutions learned are tied to a fixed discretization. In particular, there is no guarantee that neural networks generalize to other discretizations, and they often perform poorly when interpolated to higher resolutions~\citep{nature24nos}.

To address these limitations, a new class of machine learning models, known as \emph{neural operators}, was proposed~\citep{li21fno,kovachki23jmlr,Berner2025Principles}.
While standard neural networks learn mappings between fixed-size discretizations, i.e., finite-dimensional spaces, neural operators can directly learn mappings between functions, i.e., infinite-dimensional spaces~\citep{kovachki2024operator,Berner2025Principles}.
They are built from first principles to ensure a \emph{discretization convergence} property: a neural operator, with a fixed set of parameters, can be applied to input functions given at any discretization. Specifically, for an input function given at various discretizations, the outputs differ only by a discretization error that converges to zero as the discretization is refined. 

Neural operators are ideally suited for solving scientific problems, such as PDEs, where the solution is naturally formulated as a map between infinite-dimensional function spaces~\citep{kovachki23jmlr}. However, they present a learning curve to newcomers, and it is non-trivial to correctly implement neural operators that preserve their ability to operate on any input/output resolutions while maintaining their discretization convergence. %
This library offers state-of-the-art, batteries-included implementations of neural operator models and building blocks, as well as all the tools necessary for training and inference, allowing practitioners to seamlessly use it in their applications and combine it with existing PyTorch codebases.

\begin{figure}[t]
    \centering
    \includegraphics[width=\textwidth]{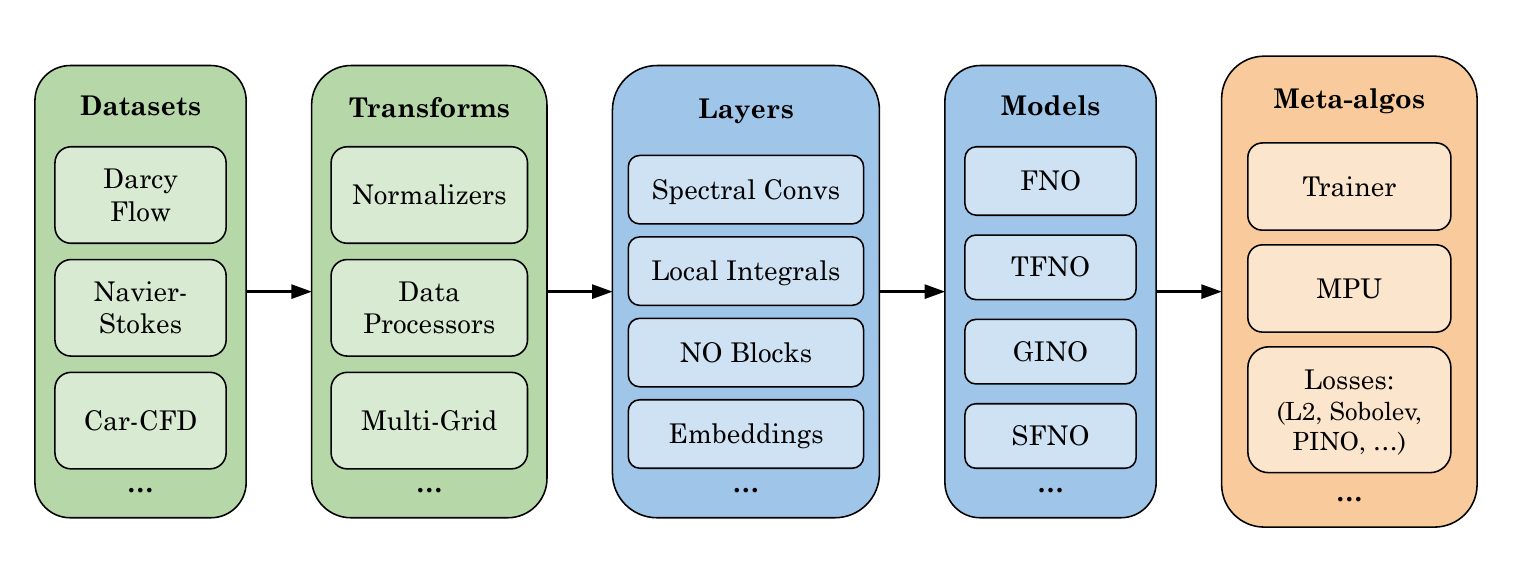}
    \vspace{-1.5em}
    \caption{\textbf{Overview of functionalities.} The \textsc{NeuralOperator} library provides the full stack required to train and deploy neural operators, including \textbf{datasets} and \textbf{data loaders}, \textbf{core layers and building blocks}, \textbf{neural operator models}, \textbf{losses} and \textbf{training and efficiency methods}. }
    \label{fig:neuralop_overview}
    \vspace{2pt}
\end{figure}

\section{The \textsc{NeuralOperator} Library}

\textsc{NeuralOperator} is an official PyTorch Ecosystem\footnote{\url{https://landscape.pytorch.org/}}\! project, open-sourced under MIT license\footnote{\textsc{NeuralOperator} is hosted at \url{https://github.com/neuraloperator/neuraloperator}.}. The \textsc{NeuralOperator} library builds on the following \textbf{guiding principles}:
\begin{itemize}[leftmargin=1.2em]
    \itemsep0em 
    \item \textbf{Resolution-agnostic:} As the crucial difference to existing frameworks, modules in \textsc{NeuralOperator}, such as data loaders, architectures, and loss functions, should be applicable to functions at various discretization. As an example, our implementation of the Fourier Neural Operator (FNO)~\citep{li21fno,duruisseaux2025FNOGuide}, just like the theoretical model, can take inputs on regular grids of any resolution. Where applicable, we also support situations where the input and output resolutions of the model differ, such as super-resolution.
    
    \item \textbf{Easy-to-use and beginner-friendly:} Out of the box, \textsc{NeuralOperator} provides all necessary functionality to apply neural operators to real-world scientific machine learning problems. It contains simple interfaces to prebuilt neural operator architectures from the literature, subcomponent layers, a \texttt{Trainer} module that automates the common training procedure, and additional submodules, including data modules and common loss functions for training neural operators. The library provides example training scripts and a detailed documentation, including an API reference, a user guide, and tutorials on layers and scientific datasets, from basics to advanced customization. A comprehensive practical guide to FNOs with \textsc{NeuralOperator} is also available~\citep{duruisseaux2025FNOGuide}.

    \item \textbf{Flexible for advanced users:} The library is designed to be highly modular. It enables a fast learning curve, rapid experimentation, and configurability, while providing a simple interface to neural operators for newcomers. This is achieved by providing various layers of modularity, from end-to-end models to more involved blocks such as spherical convolutions or tensor factorizations and algorithms such as multi-grid domain decomposition and incremental learning. We aim for the package to grow along with the field and continue providing state-of-the-art architectures, and welcome contributions from the community.
    
    \item \textbf{Reliable:} The library is designed with reliability in mind. All core functionalities are covered by unit tests, and new additions are validated with an automated CI/CD pipeline that includes tests from the function level up to end-to-end model testing. Moreover, it is documented both in-line in the source code and on an accompanying documentation website and API reference that are automatically built in CI/CD. 
\end{itemize}

\paragraph{Neural Operator Building Blocks.} 
The library offers a variety of prebuilt operator layers, implemented following the \texttt{torch.nn.Module} convention, that can be combined to build any existing or new neural operator model. These layers can be broken down by category into integral transforms~\citep{li2020gno,li21fno,kovachki23jmlr,bonev23sfno}, pointwise operators, positional embeddings, multi-layer blocks, and extra functionality (padding, normalization, interpolation, etc.). 
 
\paragraph{Neural Operator Architectures.}

\textsc{NeuralOperator} provides state-of-the-art neural operator architectures and training recipes. The core building block of neural operators is the integral transform, a learnable map between two functions supplied at any two meshes. We provide several implementations for different neural operator architectures. Using a general learnable kernel integral, \emph{Graph Neural Operators} (GNOs)~\citep{li2020gno} allow learning on arbitrary geometries.
When inputs are defined over a regular grid, the kernel integral can be efficiently realized through a Fast Fourier Transform, resulting in the spectral convolution layer used in FNOs~\citep{li21fno}. We also implement \emph{Tensor FNOs} (TFNOs)~\citep{TFNO}, which use low-rank tensor decompositions. When working on a sphere, we can leverage the Spherical Fourier transform, which is implemented via the spherical harmonic transform as proposed in~\emph{Spherical FNOs} (SFNOs)~\citep{bonev23sfno}. The efficiency of FNOs and SFNOs on structured grids can be combined with GNOs or optimal transport, yielding the \emph{Geometry-informed Neural Operators} (GINOs)~\citep{li23gino,linEnablingAutomaticDifferentiation2025} and Optimal Transport Neural Operators (OTNOs)~\citep{li2025geometricoperatorlearningoptimal}. Locally supported kernels to learn differential and integral operators (LocalNOs)~\citep{liu-schiaffini2024neural} can also supplement spectral convolutions. Recurrent Neural Operators (RNOs)~\citep{liu2023tipping} extend FNOs to systems with long-term temporal dependencies by introducing recurrence and memory directly on function spaces. \emph{Physics-Informed Neural Operators} (PINOs)~\citep{li2024pino,ganeshram2025fcpinohighprecisionphysicsinformed} can also be trained and finetuned leveraging derivative computation routines and other functionalities available in \textsc{NeuralOperator}.

\paragraph{Datasets.} We also provide convenient interfaces to common benchmark datasets for training operators on PDE data. The data is hosted on a public Zenodo archive
(at \url{https://zenodo.org/communities/neuraloperator}) and available for automatic download through the library's modules. This includes input-output pairs governed by Darcy's law, a 1d viscous Burger's equation~\citep{li2024pino}, 2d Navier-Stokes equations~\citep{TFNO}, as well as 3d Navier-Stokes equations over the surface of car models~\citep{umetani2017carcfd}. 

\paragraph{Trainers and meta-algorithms.}
\textsc{NeuralOperator} provides a suite of functionalities to simplify the training and deployment of neural operators. We provide a flexible \texttt{DataProcessor} to pipeline all normalization and transformations needed to convert inputs and outputs into the forms expected by the model and for computing losses. These also ensure that discretization convergence is respected in operations such as domain padding. We also provide a minimal \texttt{Trainer} that automates the standard logic of a machine-learning training loop, applying the data processor, optimizing model parameters, and tracking validation metrics throughout training. The \texttt{Trainer} is modular and highly configurable; users can provide their own models, data, and objectives, and the interface provides users with the opportunity to add custom training logic for domain-specific applications, including meta-algorithms such as incremental learning~\citep{george2024incremental}. Newcomers can directly use it on their own applications, while advanced users can directly import the neural architectures and core components and modules they need into their own workflow. 

\paragraph{Efficiency}
The library also packages a variety of tools for memory-efficient training of neural operator models. To compress the learnable parameters of an FNO model by performing tensor decomposition on spectral weights~\citep{JMLR:v20:18-277,TFNO}, an interface for tensorization is directly exposed in our implementation of the FNO model and spectral convolution layers. Additionally, the \texttt{Trainer} includes a native option for quantization via mixed-precision training~\citep{tu2024mixedprecision}.
Last, training of neural operators can be done incrementally~\citep{george2024incremental}, and distributed using the built-in multi-grid domain decomposition~\citep{TFNO}. These features further expand the library's capabilities and accessibility by enabling the learning of larger-scale operators on GPU. 

\section{Conclusion}
\textsc{NeuralOperator} provides state-of-the-art neural operator architectures and associated functionality in a modular, robust, and well-documented package. Built on top of PyTorch, its simple interfaces and modular components offer a gentle learning curve for new users while remaining highly extensible for conducting real-world experiments with neural operator models. It aims to democratize neural operators for scientific applications, grow alongside the field, and provide the latest architectures and layers as the state-of-the-art progresses.

\section*{Acknowledgement} %
David Pitt is supported by the Schmidt Scholars in Software Engineering program. Anima Anandkumar is supported in part by the Bren endowed chair, ONR (MURI grant N00014-23-1-2654), and the AI2050 senior fellow program at Schmidt Sciences.

\vskip 0.2in
\bibliography{references}

\end{document}